\documentclass[english]{article}
\usepackage[T1]{fontenc}
\usepackage[latin9]{inputenc}
\usepackage{geometry}
\usepackage{babel}
\usepackage{float}
\usepackage{units}
\usepackage{textcomp}
\usepackage{url}
\usepackage{graphicx}
\usepackage[numbers]{natbib}
\usepackage[unicode=true]
 {hyperref}

\makeatletter

\providecommand{\tabularnewline}{\\}

\date{}

\@ifundefined{showcaptionsetup}{}{%
 \PassOptionsToPackage{caption=false}{subfig}}
\usepackage{subfig}
\makeatother

\begin{document}

\title{STAR-RT: Visual attention for real-time video game playing}

\author{Iuliia Kotseruba\thanks{yulia\_k@eecs.yorku.ca}\and{John K. Tsotsos\footnote{tsotsos@eecs.yorku.ca}}}
\maketitle
\begin{abstract}
In this paper we present STAR-RT - the first working prototype of
Selective Tuning Attention Reference (STAR) model and Cognitive Programs
(CPs). The Selective Tuning (ST) model received substantial support
through psychological and neurophysiological experiments. The STAR
framework expands ST and applies it to practical visual tasks. In
order to do so, similarly to many cognitive architectures, STAR combines
the visual hierarchy (based on ST) with the executive controller,
working and short-term memory components and fixation controller.
CPs in turn enable the communication among all these elements for
visual task execution. To test the relevance of the system in a realistic
context, we implemented the necessary components of STAR and designed
CPs for playing two closed-source video games - Canabalt\footnote{\url{http://adamatomic.com/canabalt/}}
and Robot Unicorn Attack\footnote{\url{http://www.adultswim.com/games/web/robot-unicorn-attack}}.
Since both games run in a browser window, our algorithm has the same
amount of information and the same amount of time to react to the
events on the screen as a human player would. STAR-RT plays both games
in real time using only visual input and achieves scores comparable
to human expert players. It thus provides an existence proof for the
utility of the particular CP structure and primitives used and the
potential for continued experimentation and verification of their
utility in broader scenarios.
\end{abstract}

\section{Introduction}

Research in computer vision and related disciplines over the last
several decades has led to a rapidly growing number of algorithms
for vision-based tasks and models of visual attention. This is hardly
surprising given the demand for such research for robotics and cognitive
systems, and availability of computational resources. However, the
problem of controlling computer vision algorithms and making them
useful for general reasoning or motor control remains largely unsolved. 

The Selective Tuning Attention Reference (STAR) \citep{Tsotsos2014}
model proposes a biologically plausible, general purpose, model of
vision. It is a theoretical framework designed to control low-level
visual processing based on current task demands. The two major parts
of STAR are the Selective Tuning (ST) \citep{Tsotsos1995,Tsotsos2011}
model of visual attention and Cognitive Programs (CPs) \citep{Tsotsos2014}.
ST is a model of visual attention, whose predictions are strongly
supported by experimental data\footnote{Readers interested in the biological plausibility and experimental
verification of ST may consult the following sources \citep{tsotsos2016visual,tsotsos2016focus,rothenstein2014attentional,Tsotsos2011,itti2005neurobiology,corbetta2002control} }. Its complex hierarchical system mimics human vision and allows for
both top-down (task driven) and bottom-up (data driven) processes
to influence visual processing. The purpose of Cognitive Programs
is to control the execution of ST by modifying the way it treats inputs
based on the current visual task and directing outputs to appropriate
parts of the framework. The Cognitive Programs paper \citep{Tsotsos2014}
is published in the journal's Hypothesis category and thus is explicitly
set up for experimental testing.

In this paper we describe STAR-RT, a test of the STAR framework on
the domain of real-time video games (\href{http://adamatomic.com/canabalt/}{Canabalt}
and \href{http://www.adultswim.com/games/web/robot-unicorn-attack}{Robot Unicorn Attack}).
We chose this particular task because it requires a significant amount
of processing (e.g. search, tracking, object detection) combined with
decision-making and reasoning all to be accomplished within real-time
performance constraints. As has been noted before \citep{Mnih2015},
modern video games provide a controlled and visually challenging environment
for research in vision and AI. 

The concept of Cognitive Programs extends and modernizes the theory
of Visual Routines (VRs) \citep{Ullman1984}, which postulates that
any vision task can be decomposed to a series of simpler context-independent
subtasks. For the original VRs, visual processing is described in
\citep{Ullman1984} as a two-stage process in accordance with the
influences at that time, namely Marr's theory of vision \citep{Marr1979}.
The initial base representation (primal sketch and $2\nicefrac{1}{2}$-D
sketch in Marr's terms) is a result of the bottom-up processing of
the data. It contains local descriptions of depth, orientation, color,
and motion. Various operations (visual routines) are then applied
sequentially to the base representation to solve a task (e.g. line
tracing, determining spatial relationships, etc.). Attention is essential
for a functional system of visual routines. For example, attentional
mechanisms allow the application of the same routines to different
locations, limit the processing to a small region, maintain a list
of locations where the focus of attention should move next and a list
of already attended locations to prevent cyclic behavior.

Despite many changes in our understanding of vision and attention
since the original publication of the theory in 1984, the idea of
subtask sequencing in visual tasks is still supported by modern neurophysiological
and psychophysiological studies. For example, Roelfsema found neural
correlates for visual routines in the cerebral cortex \citep{Roelfsema2005},
and Yi and Ballard \citep{Yi1995} identified visual routines for
pouring coffee and making a peanut butter sandwich by analyzing 3D
eye-tracking data from human subjects. In a study by Hayhoe \citep{Hayhoe2000}
the ``change blindness'' phenomenon shows that visual system represents
only the information necessary for the immediate visual tasks, again
hinting at the modular structure of vision. 

The original publication on visual routines contains a few illustrative
examples but leaves out the technical details on assembly, execution,
and storage of visual routines. The gaps in theory were explored and
resolved to some extent by various projects applying visual routines
in a variety of domains: playing video games (Pengi \citep{Agre1987}
and Sonja \citep{Chapman1990}), simulated driving \citep{Garbis1998},
graph reasoning (SKETCHY \citep{Pisan1995}), active vision \citep{Rao1995},
visuospatial reasoning (Jeeves \citep{Horswill1995}), human-robot
interaction \citep{Rao1998} and robotics (AV-Shell \citep{Fayman1996},
AIBO \citep{Halelamien2004} and trash-collecting mobile robot \citep{PeterBonasso1997}).
It should be noted that although low-level vision and attention played
a major role in the original formulation of visual routines, it was
implemented only in a few projects. For example, Pengi and Sonja obtain
data from the game engine, omitting the low-level vision. In \citep{Rao1998}
attention state contains a current object of interest, its attributes,
its local context and also a history of previous attention states.
However, there is no fovea or explicitly defined region of interest
around the focus of attention.

As was already mentioned, Cognitive Programs retain the main premise
of Visual Routines theory but introduce more precise formulations
of the system components and interactions between them. Like visual
routines, CPs serve as means for attentional modulation of the visual
system, however the range of possible attentional functions is significantly
expanded. In addition to region of interest selection for gaze change,
it includes top-down priming, feedback processing, suppressive surround
and potentially even modulation of the operating characteristics of
individual neurons. While the original formulation of VRs implied
some memory operations (e.g. indexing) and decision-making abilities,
CPs explicitly include read-write memory access as elemental operations
and can also represent decision-making and setting control signals.
Overall, CPs can be seen as a fine-grained version of VRs. A more
detailed account of Visual Routines theory as the precursor of Cognitive
Programs can be found in \citep{Kruijne2011,Tsotsos2014}.

In addition to designing and implementing CPs, our work on STAR-RT
combines findings from several areas of research such as computer
vision, visual attention, GPU programming and AI. Thus, our model
allows us to integrate noisy low-level processing with symbolic reasoning,
which is essential for proper implementation of AI. While current
research in computer vision has made significant progress on solving
concrete problems like object detection, classification, and tracking,
these are generally used for simple applications, such as tagging
photos or surveillance. Similarly, models of visual attention explain
separate stages of visual processing and often leave out the connection
to the higher order processes. 

To date the most progress on combining these areas has been made in
the field of cognitive modeling. For example, a recently introduced
cognitive architecture named ARCADIA \citep{Bello2016} studies attention
as a central mechanism for perception and cognition. ARCADIA allows
both bottom-up and top-down influences to affect visual processing.
It also implements several types of visual memory (iconic, short-term
and long-term) and gaze control. ARCADIA can accurately model human
behavior in several psychophysical experiments: change blindness,
multiple object tracking, and inattentional blindness. 

In general, given the importance of perception for general cognition,
very few cognitive architectures give it a proper treatment. We found
that out of 31 established cognitive architectures listed on \url{cogarch.org}
and \url{bicasociety.org}, two-thirds omit perception and focus more
on high-level cognition (e.g. planning, reasoning, language comprehension,
etc.), which relies on symbolic manipulation and complex knowledge
structures. These are needed for categorical reasoning, playing turn-based
games (ticktacktoe, chess, etc.), solving puzzles, etc. Perception
is often replaced by symbolic input which can be used directly for
reasoning (CAPS \citep{Just2007}, Metacat \citep{Marshall2006},
Teton \citep{VanLehn1989}, IMPRINT \citep{Mitchell2003a}). It is
also common to simulate high-level perception by specifying visual
properties (e.g. color, shape, coordinates) of the objects in the
scene (e.g. ACT-R \citep{Nyamsuren2013a}, Soar \citep{Wintermute2009c},
EPIC \citep{Meyer1997}, Homer \citep{Vere1990}). Essentially, treatment
of perception as an independent module assumes that switching from
a simulated domain to a real environment can be done by replacing
one ``black box'' with a more sophisticated one. 

The architectures that are working with real sensory data provide
evidence that perception should be more tightly integrated with the
rest of the system. Some steps in this direction are being undertaken
already as there is a demand for more practical applications for cognitive
systems, e.g. robot navigation or autonomous driving. For instance,
the robotic architectures such as RCS \citep{Albus1995} and ADAPT
\citep{Benjamin2013} process high volumes of sensory data in real
time for autonomous navigation. Even when not explicitly stated, elements
of visual attention (selection of areas of interest or feature selection)
are often employed to reduce the computational load. A lot of effort
is also spent on filtering the sensor data and resolving the issues
caused by the noise coming both from sensors and changes in the environment.
Because in such systems the perceptual component is dominant, they
are restricted to the tasks that do not require a lot of symbolic
processing (e.g. navigation, object recognition, tracking, etc.) and
are not easily extended to other domains. More biologically sound
architectures like BECCA \citep{Rohrer2012} and LEABRA \citep{OReilly2012}
are also limited to recognition or visual tracking, which do not require
any reasoning. Both architectures are yet to be applied to realistic
data. Currently, LEABRA recognizes 100 categories of objects from
synthetic images with plain backgrounds and BECCA works on synthetic
$144\times144$ images or in high contrast visual environments. 

Finally, we would like to address our choice of video games as a testbed
for the STAR framework and briefly discuss relevant research on game
AI. In a sense, interaction with computer games is analogous to the
interaction with the physical world, albeit in a simplified and controlled
form. Video games of various genres make a challenging setting for
research in general AI \citep{laird2002research}. Currently, game
AI focuses on both designing artificial opponents to the human players
and imitating the human style of playing using classic AI tools (path-planning,
finite state machines) as well as machine learning techniques. 

We consider results of game AI challenges indicative of the best techniques
for achieving high scores rather than due to the theoretical novelty.
Since the mid-2000s, game AI challenges have become a noticeable part
of the AI research, many of which are run by universities and as part
of conferences. For example, the IEEE Conference on Computational
Intelligence and Games hosted six game AI competitions in 2015. 

Most AI research is currently conducted using game simulators, hence
the problem of uncertain and noisy perceptual information is rarely
addressed. There are few exceptions such as the Angry Birds AI competition\footnote{\url{https://aibirds.org/}},
where locations of various objects are found using color segmentation.
However, both vision and physics modules are included with the competition
software and participants are not expected to improve them. Generally,
in AI challenges the information about the world is complete and correct. 

Super Mario Bros. as one of the most representative platform games,
deserves a detailed discussion. As in most other platform games, in
Super Mario, the score depends both on the time it takes to clear
a level and bonuses collected in the process. Thus, playing the game
can be viewed as finding an optimal path through the environment.
This is a typical problem for classic algorithms such as A{*} search
\citep{P.E.1968}. In this case, the world-state is defined by the
state of Mario, locations of his enemies, bonus items and immobile
objects. Possible next states are determined by the actions (e.g.
jumping, moving, firing, etc.), and can be computed precisely by running
the physics engine to simulate the next step. The costs of actions
are assigned using heuristic, e.g. actions leading to death are heavily
penalized. Super Mario game turned out to be well suited for this
technique. As a result, the A{*}-based agents achieved the highest
scores for several years in a row in the MarioAI\footnote{\url{www.marioai.org}}
competition \citep{Togelius2012}. For comparison, a purely rule-based
agent for Super Mario performed at $\sim75\%$ of the top score achieved
by the A{*}-based agent but was more than 100 times faster. On the
other hand, in the Angry Birds AI 2015 competition, both finalists
were heuristic agents \citep{Borovicka2014,Ianni2015}. 

Solutions based on machine learning algorithms performed significantly
worse than agents based on heuristics, classic AI (Finite State Automata,
behavior trees, A{*}) and combinations of the two. However, machine
learning, in particular, deep neural networks, can be successfully
applied to playing multiple games as demonstrated by Mnih et al. \citep{Mnih2015}.
Out of 61 Atari games, their system could play more than half better
than the human experts. While it is impressive that control commands
for a wide range of game genres can be learned from the screenshots,
this model provides little insight into the cognitive processes that
led to its performance. Even though some hidden layers of the network
can be associated with particular parts of certain games, it does
not explain, for instance, why Pinball is the easiest game to play
(4500\% better than human) and Montezuma's Revenge is the hardest
(cannot be played at all by the model). 

The rest of this paper discusses the STAR framework and Cognitive
Programs, followed by the implementation details of STAR-RT for playing
online video games. 

\section{The STAR model and Cognitive Programs}

\begin{figure}
\noindent \begin{centering}
\subfloat[{\footnotesize{}\label{fig:STAR_diagram}Diagram of the STAR model}]{\includegraphics[scale=0.35]{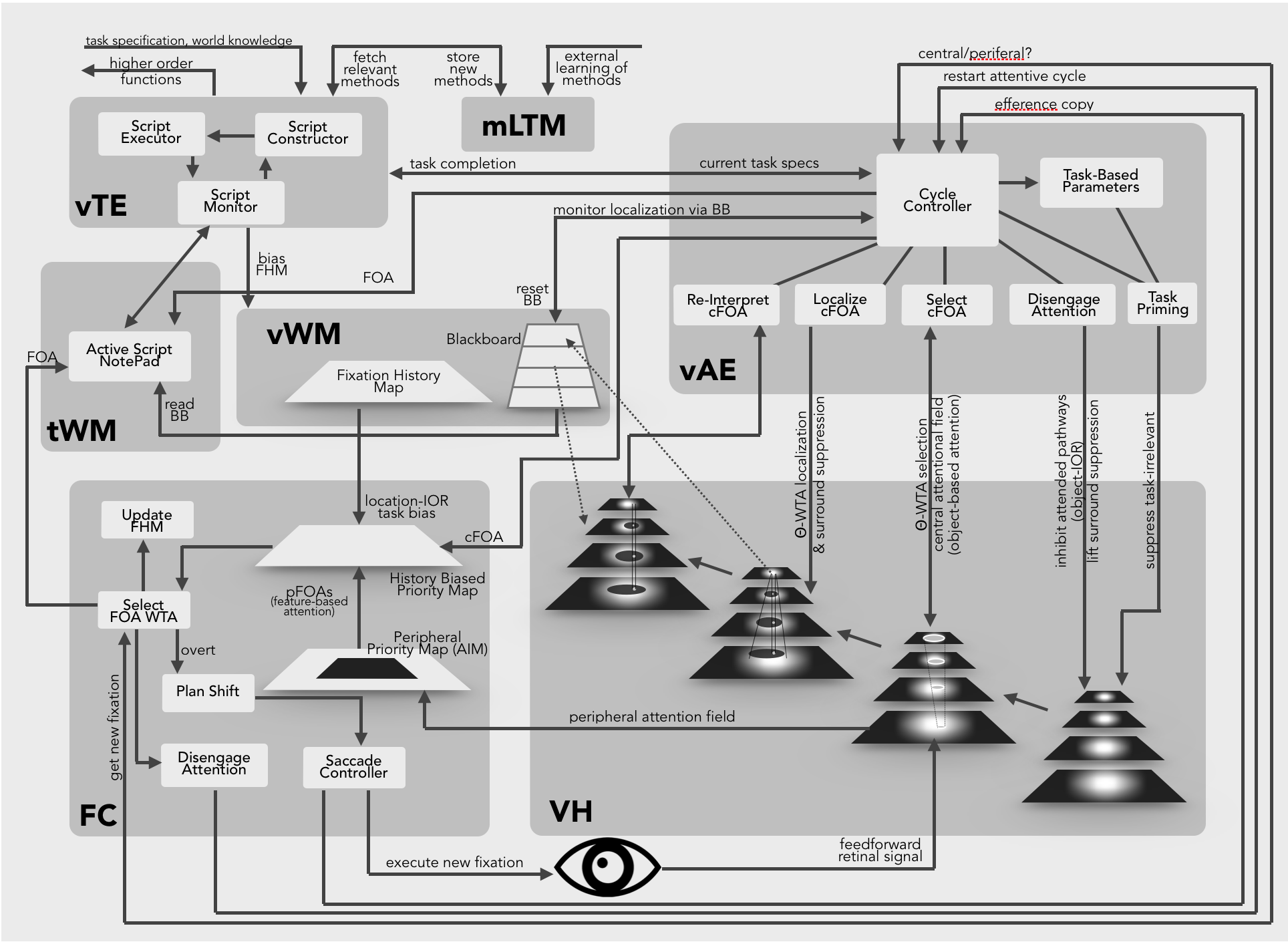}

}
\par\end{centering}
\noindent \begin{centering}
\subfloat[\label{fig:STAR-RT}{\footnotesize{} Diagram of the STAR-RT for playing
video games. Note the changes in visual hierarchy (VH) and visual
attention executive (vAE).}]{\includegraphics[scale=0.35]{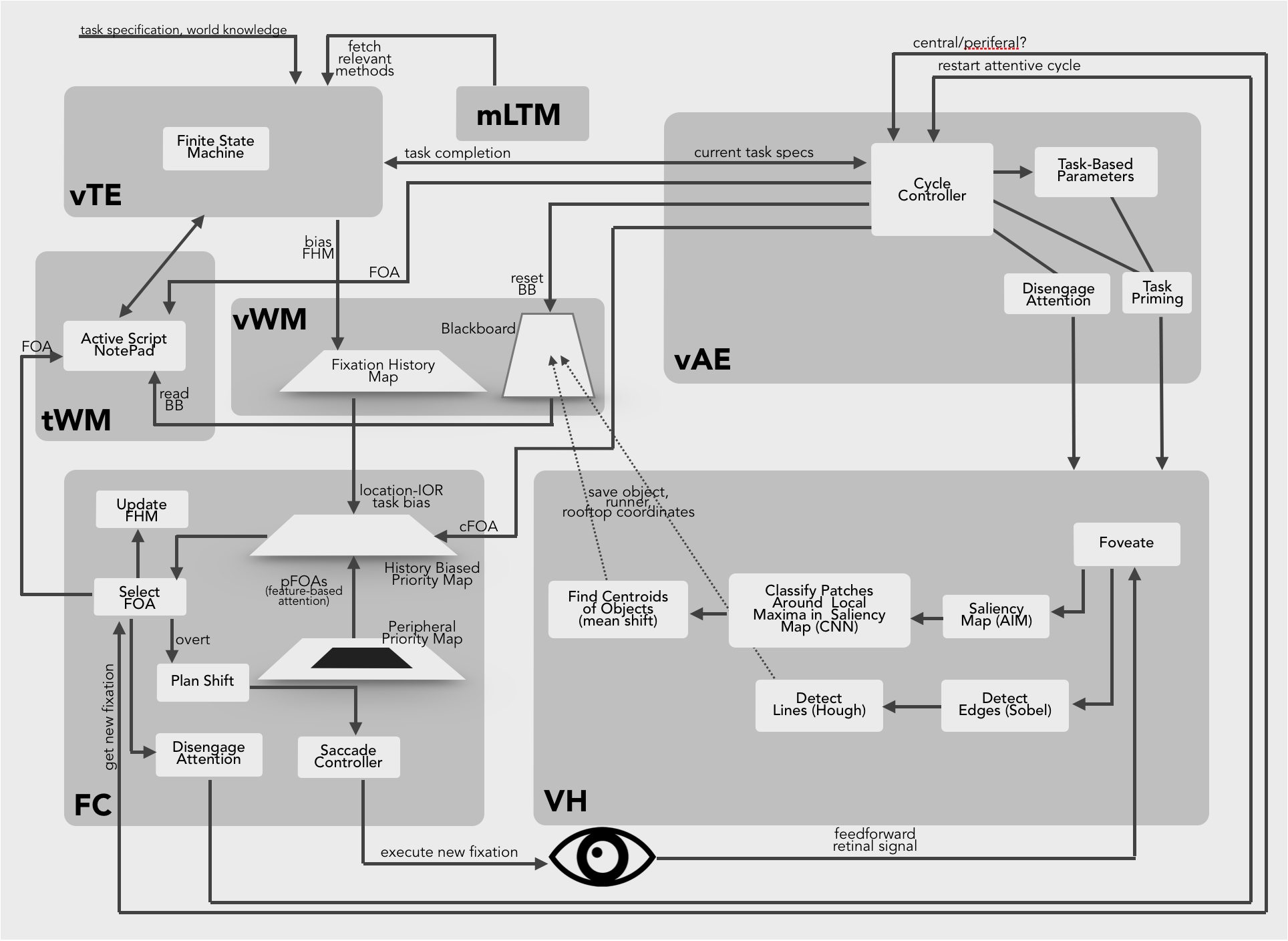}

}
\par\end{centering}
\caption{{\footnotesize{}Diagrams of STAR and STAR-RT. Gray boxes show main
elements and their internal structure. The directions of black arrows
indicate the direction of information flow within the framework.}}
\end{figure}

The STAR model is an executive controller for the Selective Tuning
(ST) model of attention. A formulation of STAR, including Selective
Tuning and Cognitive Programs (CPs) is given in \citep{Tsotsos2014},
and here we only provide a short overview of the framework (Figure
\ref{fig:STAR_diagram}). 

The Visual Hierarchy (VH) box in the diagram is based on the Selective
Tuning model of attention. ST models a hierarchy of neurons with connections
between neurons in the same layer and layers above and below. The
$\Theta$-WTA (winner-takes-all) algorithm selects features or locations
within each layer of the pyramid. Figure \ref{fig:STAR_diagram} shows
the stages of visual processing: 1) priming for the target, 2) feedforward
pass, 3) recurrent top-down localization, and 4) another feedforward
pass with suppressed units. Not all stages are necessary for every
visual task, for example discrimination and categorization would stop
after the feedforward pass 2). Furthermore, the ST hierarchy allows
priming not only for spatial locations where the target is likely
to appear but also for particular features (color, motion, etc.). 

The current focus of attention (FOA) is shown as lines within the
visual hierarchy. Selected features, current fixation and other parameters
of the VH within the FOA are called an attentional sample. Its size
and shape depends on the parameter $\Theta$ in WTA, which in turn
can be modified depending on the task. 

The remaining components in the diagram tune and control the execution
of ST. Communication between the components of STAR and parameter
setting is done via cognitive programs. CPs are composed of various
operations or other CPs and can be of two types - methods and scripts.
Methods are the blueprints of the operations with unassigned parameters.
Scripts are methods with particular values for parameters in place
and ready for execution. For example, a method for visual search would
require a target to be specified. 

Fixation Control (FC) plans the gaze change. The next fixation is
selected from the history biased priority map (HBPM) which combines
salient items from the central visual field (cFOA) and from the peripheral
visual field. cFOA is the result of processing by the visual hierarchy.
The peripheral visual field is represented by the peripheral priority
map (PPM), which is computed in a bottom-up fashion using the AIM
saliency algorithm \citep{Bruce2005}. 

There are several types of memory in STAR: visual working memory (vWM),
task working memory (tWM), and long-term memory (mLTM). 

vWM has two elements - Blackboard (BB) and Fixation History Map (FHM).
Every time attention shifts to a new location, a new attentional sample
is generated in VH and saved in blackboard which makes the attentional
sample accessible to other components. FHM keeps track of previous
fixations to avoid revisiting previously seen locations. However,
it can also be overridden if needed for the current task.

tWM saves all relevant information about the scripts in progress in
Active Script NotePad. In addition, tWM has access to the attention
sample and cFOA stored in vWM.

mLTM is an associative memory store that contains predefined methods
and the methods used in the previous tasks. The required methods are
fetched from the long-term memory using the details of the task as
indices.

It is worth noting that visual routines were presented in the original
paper as chains of functions, where the output of the every function
is fed as input into the next function along the chain. Consequently,
there was no need for the long-term storage for the visual routines
themselves, short-term memory for the intermediate results or I/O
functions. On the other hand, all practical implementations of visual
routines had some sort of temporary storage, such as registers for
saving the indexed locations (surprisingly, none related this to a
concept of working memory). However, none of the existing implementations
of visual routines explicitly defined long-term memory or equivalent
structures for storing and retrieval of elementary operations. Similarly,
long-term memory is not discussed in the psychological studies on
visual routines.

The two components which control the execution of the cognitive programs
within STAR are visual attention executive (vAE) and visual task executive
(vTE). vAE initiates and terminates each stage of VH until the vTE
sends the command that the task is finished. vTE receives a task description
from the user, selects appropriate methods from the mLTM, tunes them
into scripts and runs them using the data stored in tWM.

The aforementioned elements of the STAR framework (with the exception
of ST) were published as a hypothesis in \citep{Tsotsos2014}. In
this paper we provide a first working prototype of STAR and CPs to
determine the feasibility of the overall concept for a challenging
visual task. In the following sections we describe an implementation
of the STAR-RT framework and results of its application to the video
game domain.

\section{STAR-RT and Cognitive Programs for playing video games}

\subsubsection*{Controls and gameplay}

The choice of games for a test of our extension to visual routines
was very important. Some similar past projects (e.g. Pengo and Sonja
\citep{Chapman1990,Agre1987}) used the exploration type of games
that required a player to interact with various objects on the screen
and solve puzzles. The demand for interaction with multiple objects
shifted the focus from vision and attention to developing complex
gameplay strategies and resolving semantic problems. STAR primarily
addresses visual behavior and testing its hypotheses would be best
accomplished by fast-paced, visually complex but conceptually simple
games. As a result, we selected two popular browser games for testing
STAR-RT: Canabalt (2009) and its clone, Robot Unicorn Attack (2010).
Both are 2D side-scrolling endless runner games featuring an infinite,
procedurally generated, environment. Although the games differ in
minor details, the objective in both is to run as far as possible,
while jumping over the gaps and avoiding obstacles. Since the speed
of scrolling gradually increases, reaction time and attention are
emphasized over planning. The player controls the running man in Canabalt
by pressing X key to jump, in Robot Unicorn Attack, Z and X to jump
and dash respectively. 

\begin{figure}[t]
\noindent \begin{centering}
\subfloat[rocks and debris after the robotic drill drops down]{\includegraphics[scale=0.15]{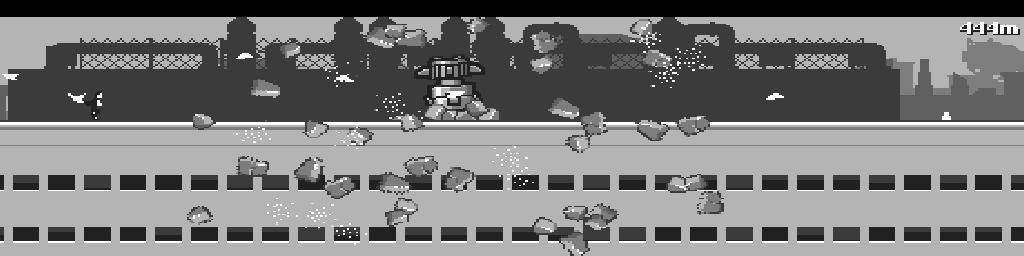}

}\subfloat[flocks of birds]{\includegraphics[scale=0.15]{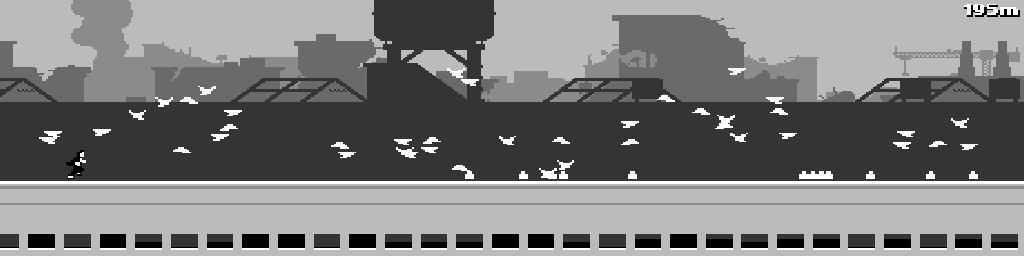}

}
\par\end{centering}
\noindent \begin{centering}
\subfloat[shards of glass]{\includegraphics[scale=0.15]{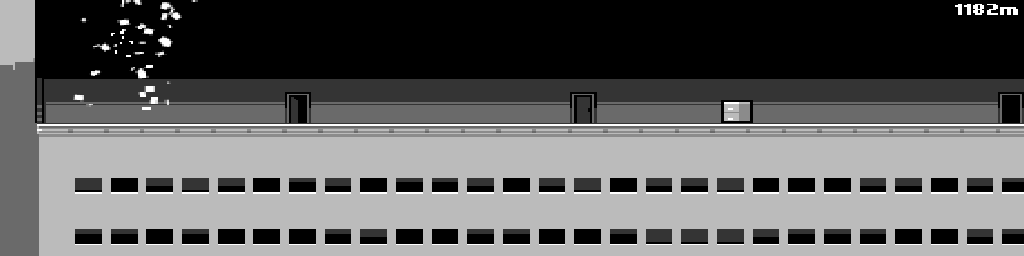}

}\subfloat[collapsing building]{\includegraphics[scale=0.15]{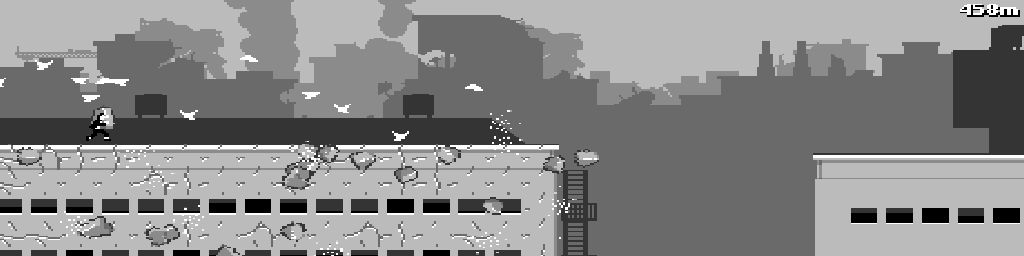}

}
\par\end{centering}
\caption{{\footnotesize{}\label{fig:canabalt_visual_distractors}Screenshots
from Canabalt demonstrating various visual distractions typical for
the game}}
\end{figure}

\begin{figure}[t]
\noindent \begin{centering}
\subfloat[{\footnotesize{}Dolphins}]{\noindent \centering{}\includegraphics[scale=0.17]{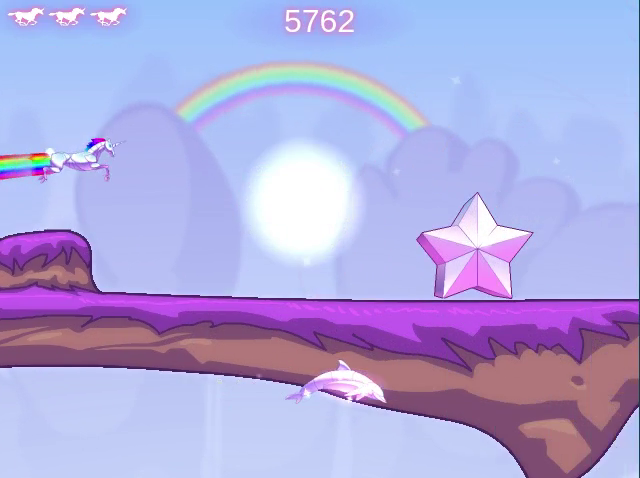}}\,\,\subfloat[{\footnotesize{}Sparkles}]{\noindent \begin{centering}
\includegraphics[scale=0.17]{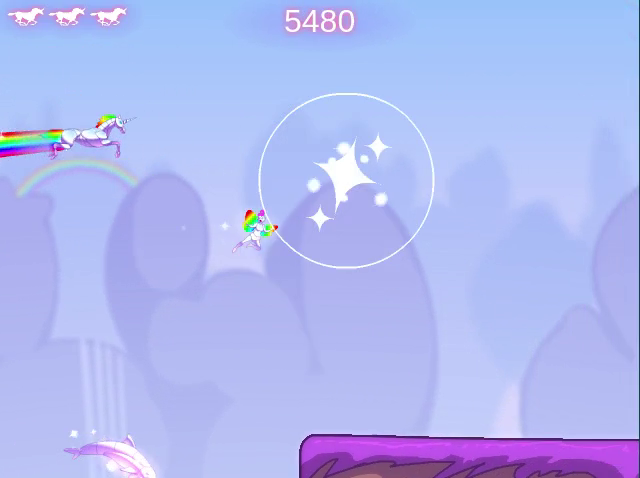}
\par\end{centering}
}\,\,\subfloat[{\footnotesize{}\label{fig:Curved-platform}Curved platform}]{\noindent \begin{centering}
\includegraphics[scale=0.17]{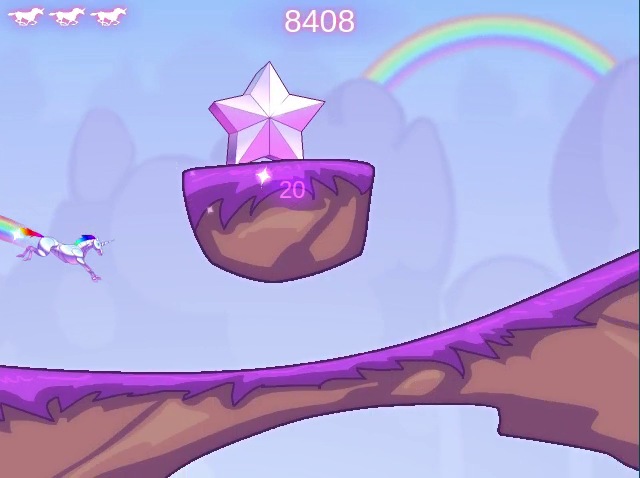}
\par\end{centering}
}
\par\end{centering}
\caption{{\footnotesize{}\label{fig:curved_platforms_robot_unicorn}Examples
of visual distractors in Robot Unicorn Attack. Dolphins are inactive
game objects that appear at the bottom of the screen. Their number
depends on how many stars have been destroyed in sequence. Sparkling
and shiny artifacts are generated whenever points are earned, e.g.
from collecting a fairy or destroying a star. Sparkles are inactive
game elements, but they usually remain on the screen for several frames
and can be distracting.}}
\end{figure}

\begin{figure*}[t]
\noindent \begin{centering}
\subfloat{\noindent \centering{}\includegraphics[scale=0.13]{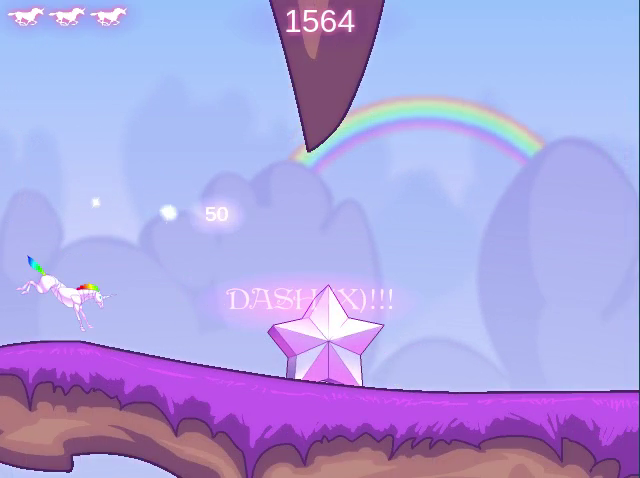}}\,\subfloat{\noindent \centering{}\includegraphics[scale=0.13]{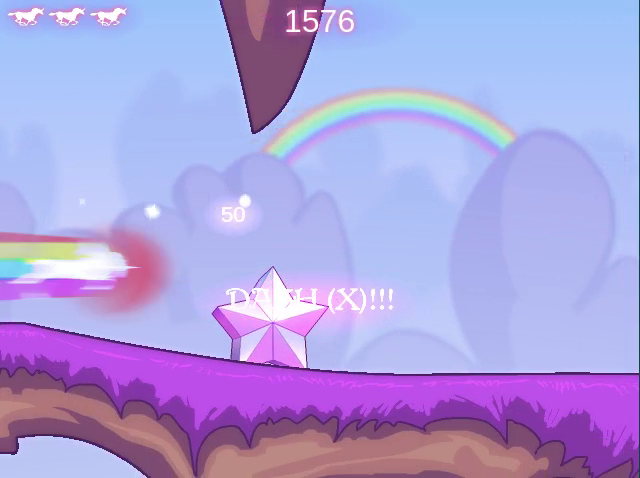}}\,\subfloat{\noindent \centering{}\includegraphics[scale=0.13]{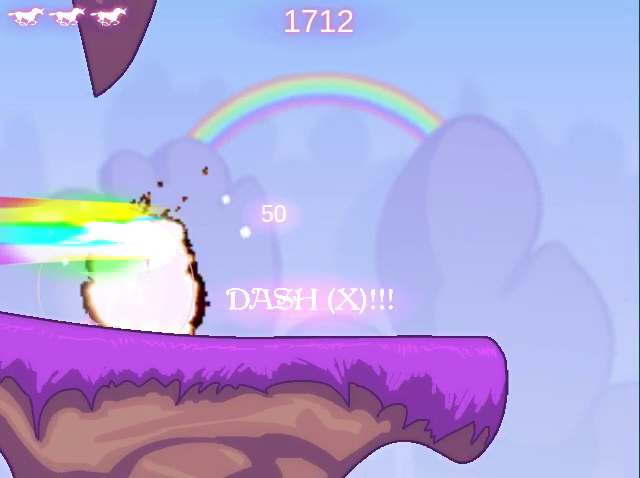}}\,\subfloat{\noindent \centering{}\includegraphics[scale=0.13]{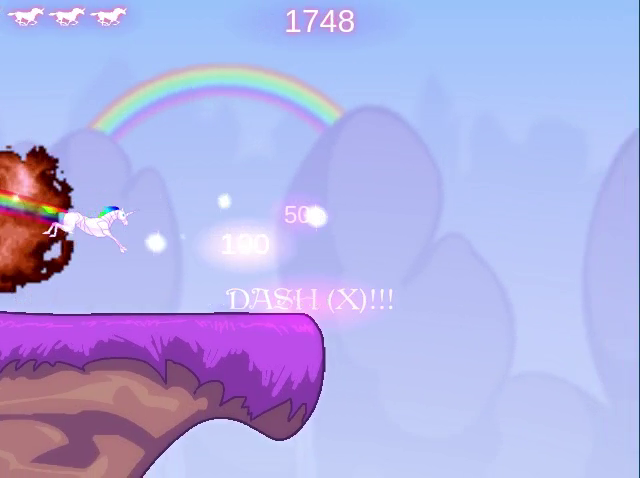}}
\par\end{centering}
\caption{{\footnotesize{}\label{fig:unicorn_dashing}Screenshots showing changes
in appearance of the unicorn when dashing through the star. Here the
unicorn is invisible behind the explosion for almost 50 frames.}}
\end{figure*}

Despite minimalist controls and gameplay, these games are not primitive
from a visual processing point of view. The smooth, detailed and endlessly
varied graphics is what makes them long-standing hits and also an
excellent testbed for models of visual attention. In Canabalt the
office worker escaping from the robot invasion encounters debris from
collapsing buildings, flocks of doves, abandoned office furniture,
shards of broken glass and robotic drills dropping from the sky (Figure
\ref{fig:canabalt_visual_distractors}). The distractors in Robot
Unicorn Attack are various sparkly artifacts when bonus points are
collected, silver dolphins and exploding stars (Figure \ref{fig:curved_platforms_robot_unicorn}
and \ref{fig:unicorn_dashing}). Curved platforms further complicate
visual processing. 

\subsubsection*{STAR-RT}

STAR-RT has no access to the game engine, therefore all information
must be obtained from the screenshots of the browser window. In order
to enable processing of high-resolution images in real time several
changes were introduced to the original formulation of STAR (Figure
\ref{fig:STAR-RT}).

Vision module output consists of the coordinates and properties of
the platforms and objects in the scene. VH is intended as a general
purpose model of vision and, naturally, it is the most computationally
expensive component of the framework. We found that implementing even
a subset of the full visual hierarchy is infeasible within our time
requirements even with most of the computation moved onto GPU. As
a result, VH currently uses only the bottom level of the pyramid and
all processing is done in a feedforward fashion. For example, recognition
of multiple objects in the scene is done in parallel on GPU instead
of serial processing with covert gaze change, and recurrent localization
is replaced with the mean shift algorithm \citep{Fukunaga1975}. Our
implementation represents low- and high-level visual processing, implements
spatial priming, has several adjustable parameters and the same connections
to the rest of the components of STAR.

The primary role of fixation control (FC) is to plan the next move
while taking into account the salient objects in the peripheral priority
map (PPM) computed using the bottom-up saliency algorithm AIM \citep{Bruce2005}. 

The visual attention executive (vAE) in STAR-RT mainly serves for
task priming and controlling the contents of the visual working memory
(vWM). Spatial priming is performed when searching for the character,
since its approximate location is known. Another instance of spatial
priming is for the objects, and in this case we use the fact that
objects are located on the top of the platform\footnote{Further examples of spatial priming in ST can be found in \citep{Tsotsos2011}}.
The vWM contains a fixation history map, which saves fixations in
the previous frames and also a blackboard (BB), where the locations
of all objects and lines detected by VH in the current frame are saved
and made available for all other elements in the diagram (they represent
an attentional sample for the current frame). 

The task working memory (tWM) temporarily stores the locations of
the objects and lines from several previous frames (currently the
number of frames is set to 10). As every new frame is processed, new
data from the blackboard is stored in the tWM overwriting the oldest
record. An external function is called to determine the speed of scrolling
and resolve any inconsistencies between the recorded coordinates of
the objects and platforms. All temporally consistent detections are
saved in the Active Script NotePad and later used for making game
decisions. 

The visual task executive (vTE) is responsible for coordinating all
these modules. As we mentioned before, it has three main functions:
1) to read the task specification and compose appropriate methods
for this task with tuned parameters, 2) to execute the script and
3) to monitor the execution of the script and modify the progress
of the script when needed. In our case the cognitive programs are
hard-coded and there is no need for composing scripts and tuning them.
Hence the script constructor is omitted. The finite-state machine
(FSM) that represents vTE in STAR-RT accomplishes two out of three
functions in the following way: the script monitor has access to the
active script notepad, which contains all information relevant for
the execution of the scripts (e.g. coordinates of objects in the current
and several past frames, variables and timer for a key press). The
vTE takes the task specification, in this instance a string specifying
the name of the game to be played. The pre-programmed methods are
then loaded from mLTM and executed step by step. The vTE also calls
several external functions to measure the speed of the game, compute
jump trajectory and determine the duration of key press. Keyboard
input is required to control the character on the screen, however,
the original formulation of STAR in \citep{Tsotsos2014} does not
specify how it interacts with motor functions. Therefore, we assume
that keyboard calls can be done through the vTE. 
\begin{figure}
\noindent \begin{centering}
\subfloat[CPs for Canabalt]{\noindent \begin{centering}
\includegraphics[scale=0.14]{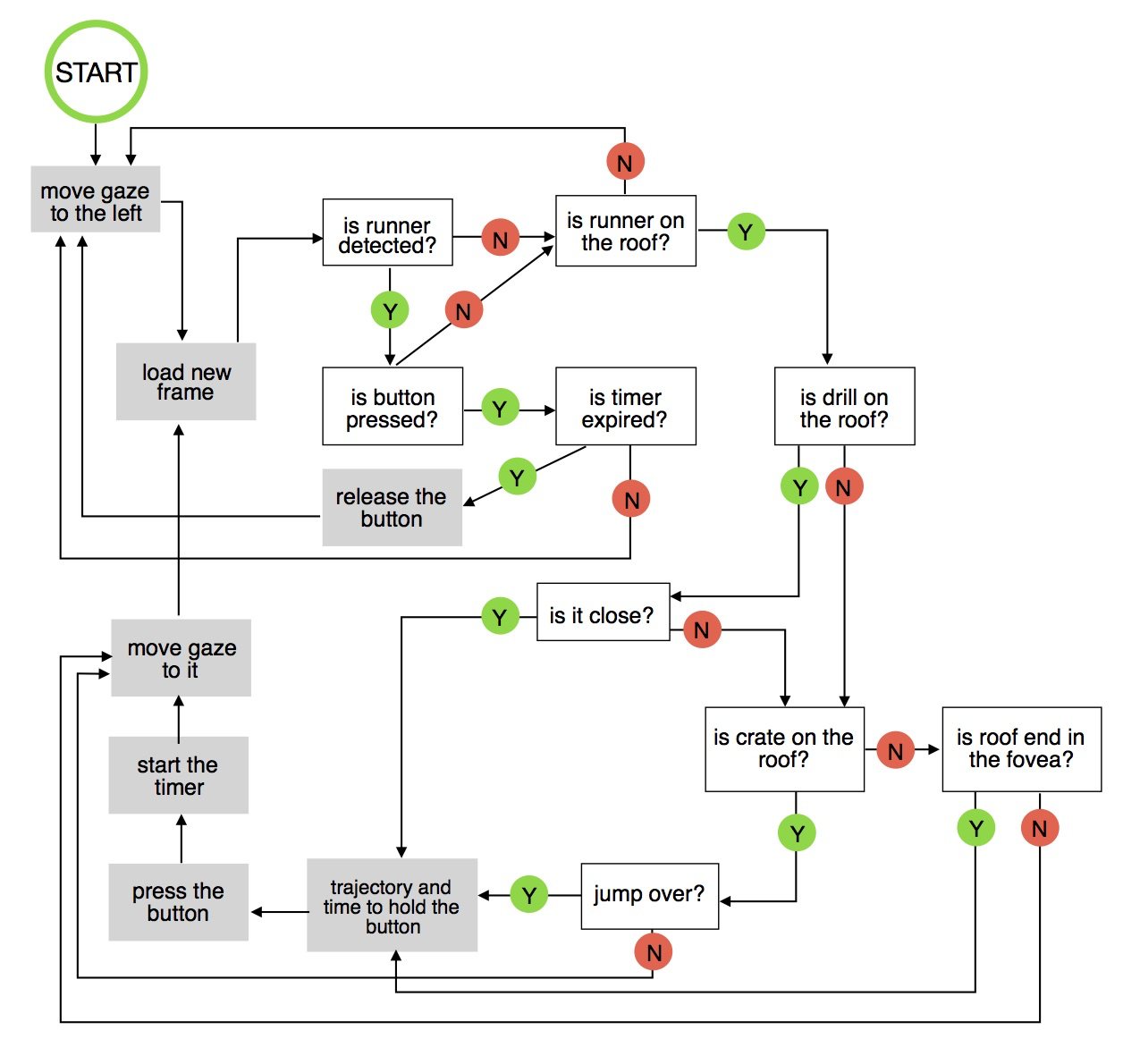}
\par\end{centering}
}\subfloat[CPs for Robot Unicorn Attack]{\noindent \begin{centering}
\includegraphics[scale=0.14]{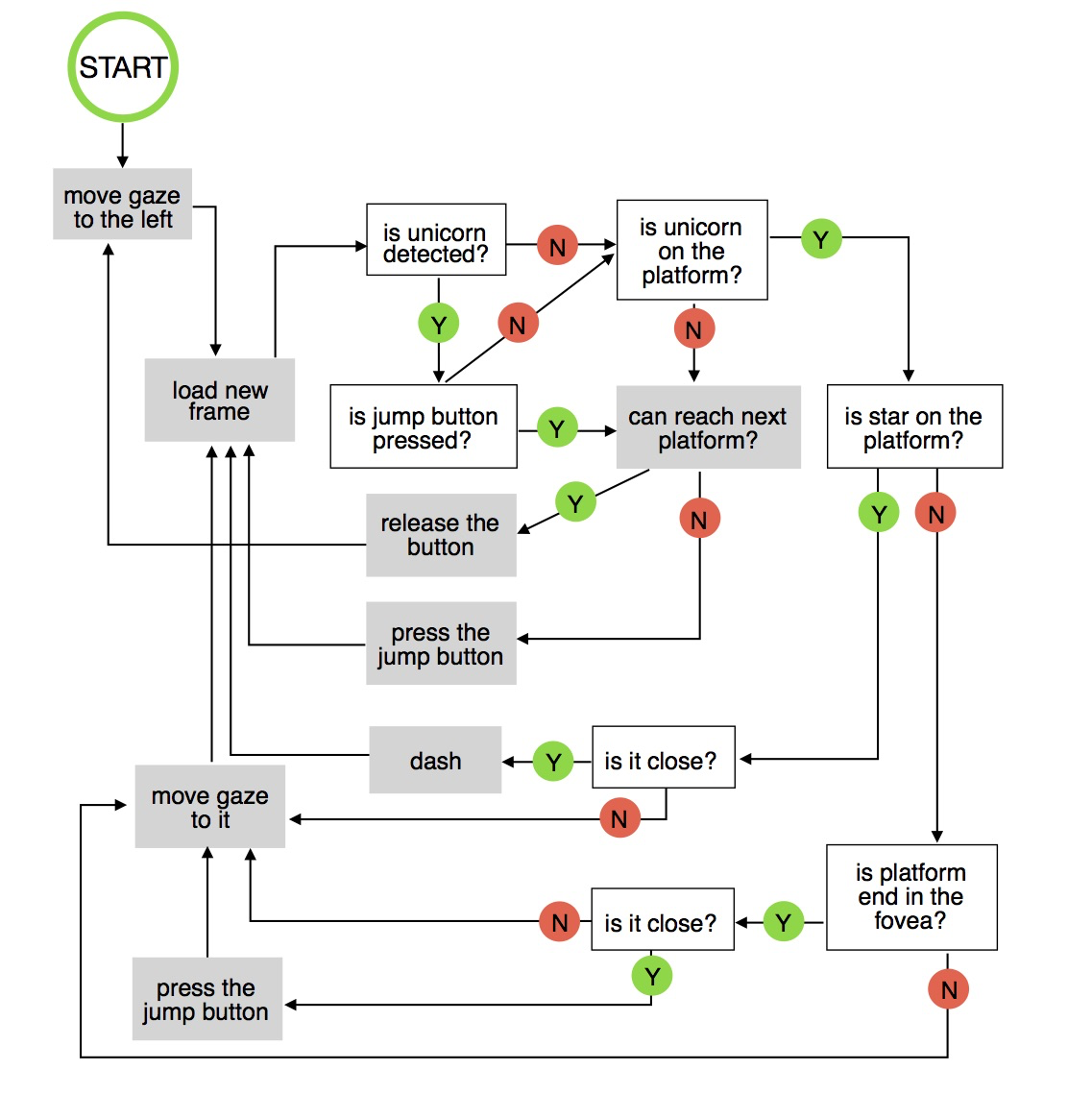}
\par\end{centering}
}
\par\end{centering}
\caption[Diagram of Cognitive Programs for the game playing task]{\label{fig:cp_diagram}{\footnotesize{}Simplified diagram of Cognitive
Programs for the game playing task. The game playing logic is implemented
as a decision tree: white boxes show methods and gray boxes show calls
to external procedures (loading a new frame, pressing a key, changing
gaze, etc.). }}
\end{figure}

\subsubsection*{Cognitive Programs for playing the games}

We formulate the task of playing a video game as a decision-making
problem, i.e. for each new frame the player must choose whether to
press/release the key given the world state (platform edges, properties
of the objects on the screen, etc). In order to develop a set of rules
we conducted a study where we asked 3 subjects\footnote{Our subjects were one female (the first author) and two males, all
members of Tsotsos lab. The first author was the only expert player
of the game, both other subjects played the game for the first time.
Participants were not paid for their time. } to play 10 sessions each of Canabalt and Robot Unicorn Attack while
tracking their eye movements with the Pupil Labs eye-tracker\footnote{\url{https://pupil-labs.com/pupil/}}.
Although both games also use audio cues to warn the player about upcoming
obstacles, our algorithm does not use it. Therefore, during the study,
all participants played the game with audio muted. Overall, 20+ minutes
of recorded data were collected for each participant. We did not continue
with a full study of eye-motion with naive participants. The study
was conducted in order to find out if any useful strategy for playing
the game can be extracted from the eye-tracking data. Our analysis
of the data was limited to observing videos with overlaid eye-tracking
data and finding patterns in the eye movements that could be helpful
in developing a game playing logic for the game. 

All subjects showed similar patterns of fixations while playing both
games. First, they looked at the character and then moved their gaze
to the right along the top of the platform. If an obstacle was found,
it was tracked until a jump could be safely made. Then players immediately
started looking for the next obstacle to the right. 

Another finding, based on interviewing the participants after the
study, was that the best strategy for jumping in Canabalt was to plan
the landing spot closer to the edge of the platform. However, this
required fine control skills, especially as the game reaches its highest
possible speed (which in Canabalt is set to 800 pixels per second
or about 286 km/h\footnote{\url{http://www.gamasutra.com/blogs/AdamSaltsman/20100929/88155/Tuning_Canabalt.php}}). 

Since both games are very fast-paced, planning for more than one step
ahead is rarely needed. In fact, when the games reach their maximum
speed, the player has approximately a quarter of a second to make
a decision, which is at the limit of human ability \citep{Amano2006}.
Nevertheless, in Canabalt there is rarely more than one obstacle present
on a single platform, therefore all decisions about the obstacles
are made in FIFO order, one at a time. This is confirmed by the human
eye-tracking data. For instance, if there is a gap followed by a box
on the next platform, the gap will be tracked first. Only once the
jump is made, the gaze would be shifted to the edge of the next platform
and then to the box on top of it. 
\begin{figure}
\noindent \begin{centering}
\includegraphics[scale=0.3]{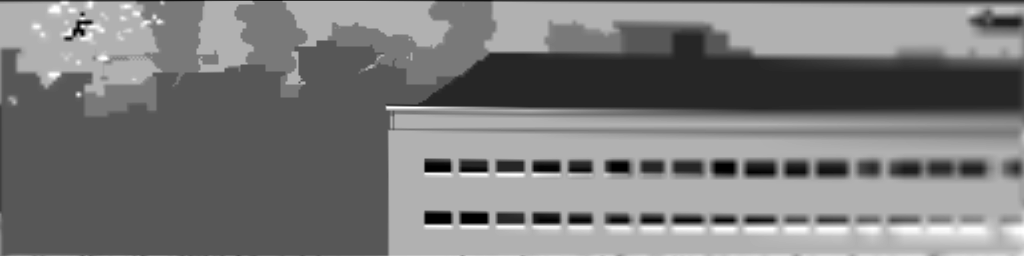}
\par\end{centering}
\caption{\label{fig:Foveated}{\footnotesize{}Foveated screenshot from Canabalt.
The fovea covers only about $\nicefrac{1}{3}$ of the frame width.
As a result, small details in peripheral areas are blurred and require
less processing.}}
\end{figure}

We applied observations made from the human player data to the design
of the Cognitive Programs for playing both games. Figure \ref{fig:cp_diagram}
shows diagrams with high-level description of methods (e.g. 'check
if the runner is on the top of the platform'). Each path within a
diagram also represents a method in the CPs terminology. In our implementation
the vTE acts as a rule-based game AI. 

The vTE controls the execution of the task based on the rules and
the information within the visual working memory and the task working
memory. For example, while playing Canabalt, to check if the character
is found in the image, the vTE examines the contents of the Active
Script NotePad. If the character is not detected and its previous
location is not within fovea, it means that the system is currently
tracking an obstacle or looking for the end of the platform. In this
case, the location of the character is assumed to be unchanged. If
no obstacles are found within the fovea and the platform extends beyond
it (``is roof end in the fovea?''), then the gaze is gradually shifted
to the right along the current platform as each next frame is loaded.
When the platform is very long, it may span several lengths of the
screen. The gaze will eventually reach the edge of the image and will
remain there until the end of the rooftop (or an obstacle) appears.
After that, the end of the rooftop (or an obstacle) will be tracked
normally and the decision to jump will be made eventually.

If the runner is found within the fovea and is not performing a jump
(``is runner on the roof?''), then the obstacle locations recorded
in the tWM are checked (``is drill/crate on the roof?'').\textit{
}If an obstacle is detected and located sufficiently close to the
character (the distance depends on the current frame rate and speed),
a decision is made whether to jump over it or not (``jump over?'').
If a jump is required, vTE calls an external function to compute the
trajectory of the jump and the duration of the key press and then
sets the timer and a flag for the key pressed in the tWM to true and
loads a new frame. 

Below we describe what actions within the STAR framework are required
to implement the elements in the diagram. More technical details and
pseudo-code can be found in \citep{Kotseruba2016}.

\textbf{START. }All parameters of the system are reset to the default
values, the visual hierarchy is primed to look for the character in
the left half of the screen, and the gaze location is moved to the
left. Since we expect the character to always be in the left $\nicefrac{1}{3}$
of the screen, we set the gaze location at ($\frac{1}{2}h,\frac{1}{3}w$),
where $w$ and $h$ are width and height of the frame. This way the
vision system will have a chance to recognize and localize a character
and most of the scene will also be visible.

\textbf{LOAD NEW FRAME.} A screenshot of the rectangular area within
the browser window is taken and resized from $320\times920$ to $256\times1024$
pixels to improve GPU processing. Next, the image is foveated using
the current gaze location (Figure \ref{fig:Foveated}). We reimplemented
the BlurredMipmapDemo from Matlab Psychophysics Toolbox \citep{Brainard1997}
on GPU. First, we build a Gaussian pyramid (without scaling), combine
levels of the pyramid so that the fovea contains pixels from the first
(not blurred) level, and copy the rest of the pixels from the different
levels of the pyramid depending on their distance from the center.
Following \citep{Strasburger2011} the fovea diameter is set to 2\textdegree .
Other parameters were set according to the conditions of the human
study, namely the distance to the monitor of 57 cm and monitor ppi
of 95.77.

\textbf{FEED-FORWARD PASS. }The feed-forward pass in the VH computes
line segments and recognizes salient objects. Localization is done
by the mean-shift algorithm \citep{Fukunaga1975}. Edges are detected
by filtering the foveated image with a $3\times3$ Sobel operator.
To find line segments we use the fast Hough transform optimized for
GPU \citep{Nugteren2011}. The line segment endpoints are computed
on CPU. As a result, for each frame at most $10$ lines with lengths
of $100$ or more pixels are detected. If the total length of gaps
is more than $15\%$ of the line length, it is discarded. 

In order to find regions of interest in the image we use a bottom-up
saliency algorithm AIM\footnote{\url{http://www-sop.inria.fr/members/Neil.Bruce/AIM.zip}},
which we ported to OpenCL. We find at most $100$ salient points\footnote{Since recognition is performed on the GPU, the structure holding local
maxima cannot be dynamically allocated. Therefore, we gathered statistic
over several thousand frames from the game and determined that majority
of frames contain only a few objects, however frames with flocks of
birds can have up to 50. In order to ensure that every local maxima
is processed, we doubled that number.} in the AIM saliency map on CPU by iteratively selecting local maxima
above the threshold (\textbf{$50\%$ }of max saliency) and inhibiting
a $20\times20$ area around them to avoid tight clustering.

\begin{figure}
\noindent \begin{centering}
\includegraphics[scale=4.5]{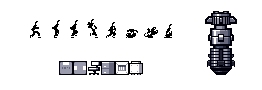}
\par\end{centering}
\caption[Objects in Canabalt]{\label{fig:canabalt_objects}{\footnotesize{}Objects in Canabalt:
top row shows several key-frames from the main character animation,
and the bottom row shows various non-lethal objects (crates and office
furniture). The robot drill is the only lethal object in the game
(shown on the right).}}
\end{figure}

Around each local maximum, $50$ normally distributed points are sampled.
These random points are the centers of $30\times30$ image patches,
which are passed to a convolutional neural network (CNN). The GPU
implementation of CNN is based on the DeepLearnToolbox \citep{Palm2012}.
Only the CNN classifier part of the code runs on GPU, while training
is done in Matlab offline on CPU. CNN must be able to distinguish
patches of 4 classes: runner, non-lethal obstacles (crates), lethal
obstacle (drill) and everything else (Figure \ref{fig:canabalt_objects}).

Each class has a considerable amount of variation. For example, the
runner\textquoteright s animation is composed of $38$ different frames,
and he sometimes could be confused with shards of glass and flocks
of birds. There are also $7$ types of non-lethal obstacles - crates,
boxes and office furniture. Robotic drills are the least varied in
appearance, but they also produce a lot of flying debris, which do
not affect the runner but occlude the view of the drill itself and
nearby objects. Since template matching and SVMs were not very good
at separating these classes, a CNN is used instead \citep{LeCun1995}. 

An architecture for the network was derived experimentally and has
4 layers - 2 convolution and 2 subsampling. The network was trained
for 400 epochs with learning rate parameter (alpha) set to 1.0 and
a batch size of 50. The training data contained 15000 samples for
``other'' class and 5000 for each of the runner/crate/drill classes.
The final network accuracy is 98\%.

Finally, we cluster all points with the same class labels using the
mean shift algorithm \citep{Fukunaga1975}. This step is needed because
AIM maxima do not necessarily correspond to the centroid of a salient
object, and also because objects larger than $30\times30$ (e.g. drills)
may produce several salient points. Clustering is performed on CPU
for each class separately with bandwidth of $20$ pixels. Clusters
containing more than half of points of \textquotedblleft other\textquotedblright{}
class are ignored. All discovered line segments and centroids of objects
are then saved in the Blackboard. 

\textbf{EXTERNAL FUNCTIONS. }External functions are called by the
vTE to eliminate the false detections for objects, find current platforms,
estimate speed, compute jump trajectory and estimate key press duration. 

The starting point is the location of the runner, since its location
is the most constrained (movement is vertical with slow drift towards
the center of the screen as the speed of the game increases, but at
most 200 pixels from the left edge). The longest edge directly below
the runner is assumed to be the current platform. If there are any
objects on the screen, they are used as additional evidence. If the
current platform does not extend beyond the right edge of the frame,
we look for lines that begin after the current platform ends and select
the top one.

Matching detected objects and finding displacements are done simultaneously.
Since many of the obstacles are visually identical, the only way to
distinguish between them is by their coordinates. We compute pairwise
displacements between all detected objects in the current and previous
frames and select globally the most consistent one (or a minimum value
if all displacements are unique). The fact that the game scrolls from
right to left is an extra constraint used to eliminate incorrect displacements.
We also check that the motion of the platforms is consistent with
the object displacements. Here, the speed is estimated using displacements
from the past $15$ frames. 
\begin{figure}
\noindent \begin{centering}
\includegraphics[scale=0.3]{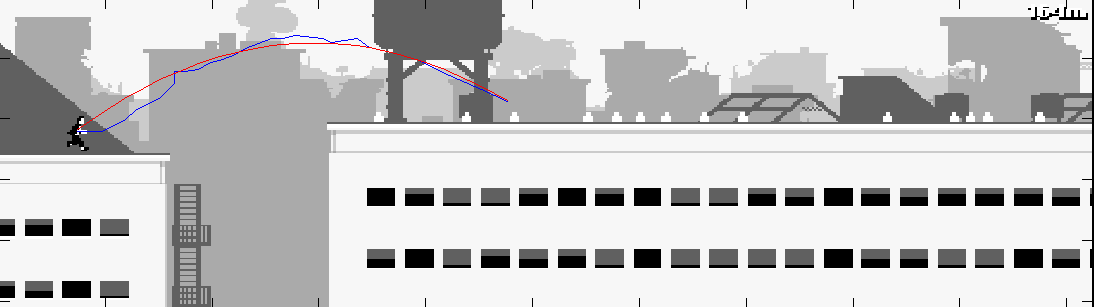}
\par\end{centering}
\noindent \begin{centering}
\includegraphics[scale=0.3]{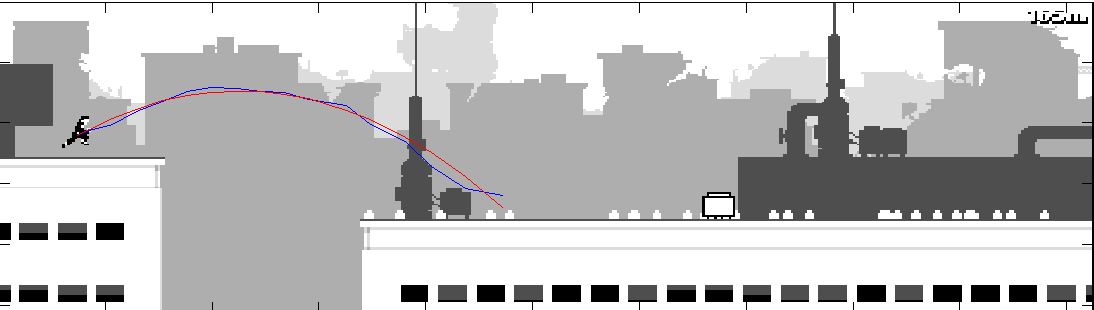}
\par\end{centering}
\caption[Fitting a parabola to a jump trajectory]{\label{fig:parabola_jump_canabalt}{\footnotesize{}Fitting a parabola
to a jump trajectory. Blue line represents the sampled trajectory
of the runner and red line shows an approximation.}}
\end{figure}

Jumping is the only action in the game and mistakes in timing or precision
affect the final game score. The optimal strategy is to plan the landing
spot closer to the edge of the next platform, because it leaves more
time to react to obstacles. The trajectory of the jump is determined
by the initial speed, the amount of time the key was pressed, the
location of the runner and the size of the obstacle. Holding the X
key maintains vertical acceleration, and jump height is proportional
to the speed. Pressing the key for more than $350$ ms has no additional
effect. 

Typically in the game AI research the future state can be obtained
by running the game engine a few steps forward, which greatly simplifies
learning the game behavior. Since both Canabalt and Robot Unicorn
Attack are closed-source, we estimate parameters of the game from
noisy data collected over multiple sessions. In Canabalt, the jump
trajectory can be well approximated by a parabola (Figure \ref{fig:parabola_jump_canabalt}),
so we use two parameters as a jump descriptor. Based on runtime statistics
on hundreds of jumps (coordinates of the runner, time stamps, ``X''
key on/off, parameters of the parabola and speed) we also find correlations
between the height of the jump and key press duration, and between
the speed and maximum height of the jump. All parameters of jump physics
are learned offline. At runtime when the runner is approaching an
obstacle, the vTE calls a function to determine whether a jump can
be made given the current speed. If yes, the key press time is returned,
if not, it usually means that the jump was planned too early. In this
case, if the runner is still far from the obstacle, the function returns
$0$ and attempts to jump later. Otherwise, if the obstacle is too
close, the biggest possible jump is performed.

\medskip{}

\begin{figure}
\noindent \begin{centering}
\includegraphics[scale=0.7]{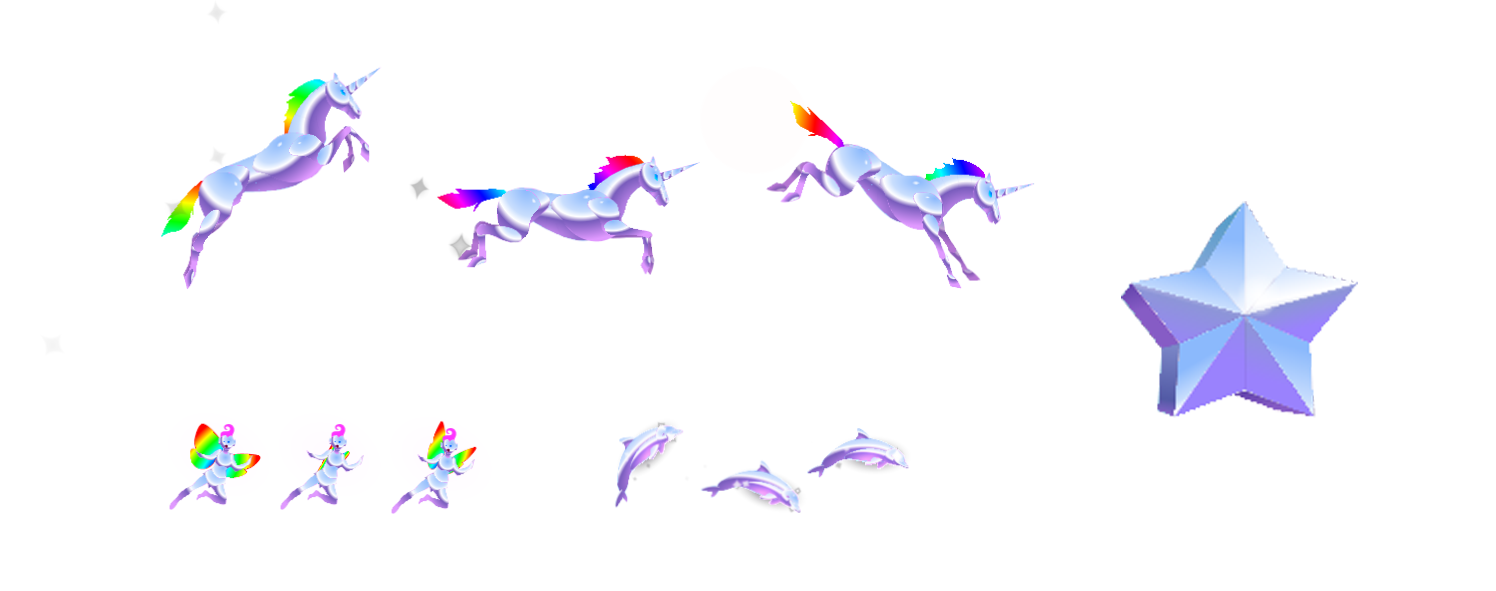}
\par\end{centering}
\caption[Objects in Robot Unicorn Attack]{\label{fig:unicorn_objects}{\footnotesize{}Objects in Robot Unicorn
Attack: top row shows several key frames from the unicorn animation.
Bottom row shows fairies and dolphins (inactive game elements). The
only lethal game element is the star, shown on the left. }}
\end{figure}

\begin{figure}
\noindent \begin{centering}
\subfloat[\label{fig:unicorn_DP}]{\noindent \begin{centering}
\includegraphics[scale=0.27]{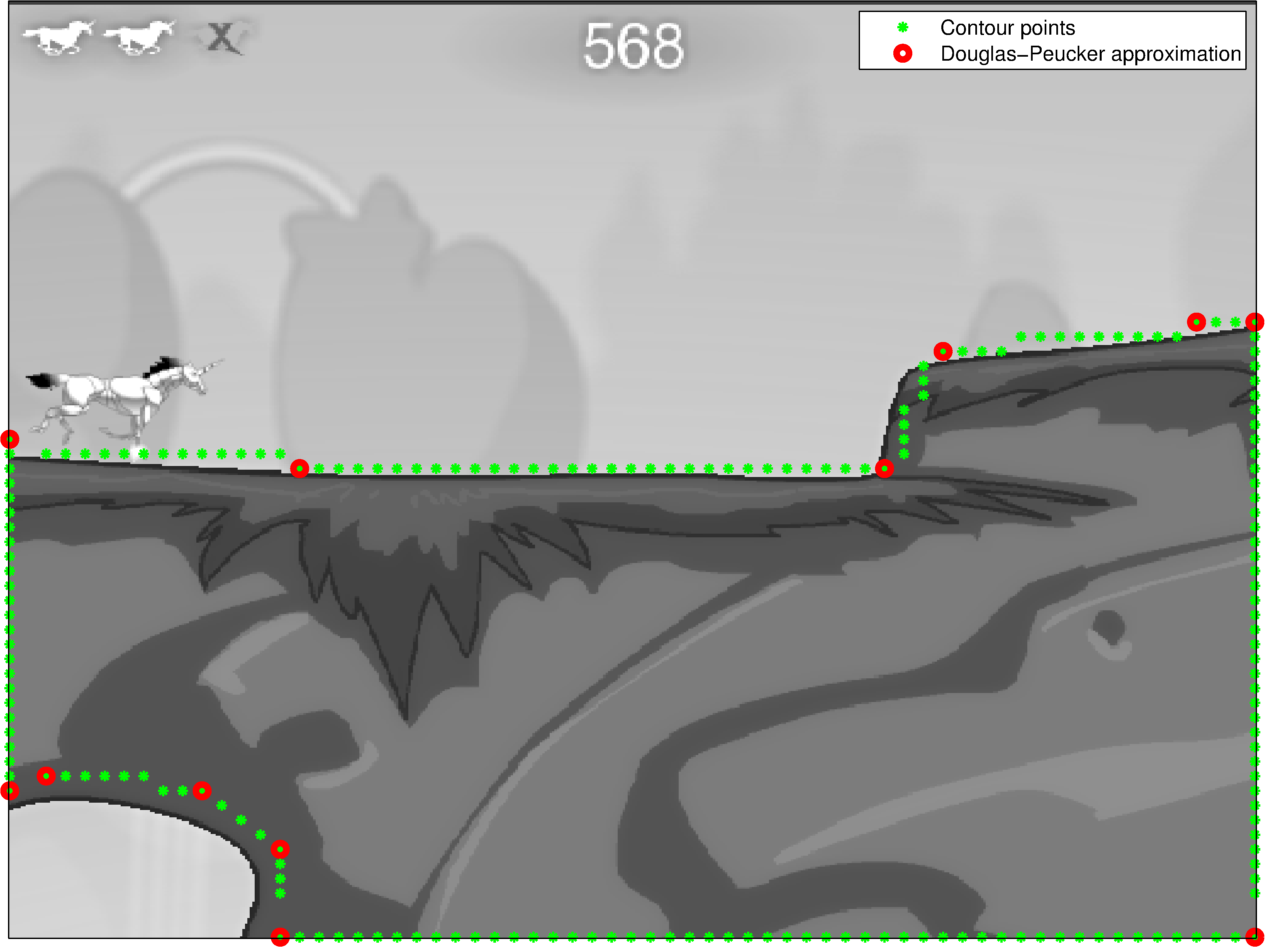}
\par\end{centering}
}\subfloat[\label{fig:unicorn_platform_top}]{\noindent \begin{centering}
\includegraphics[scale=0.27]{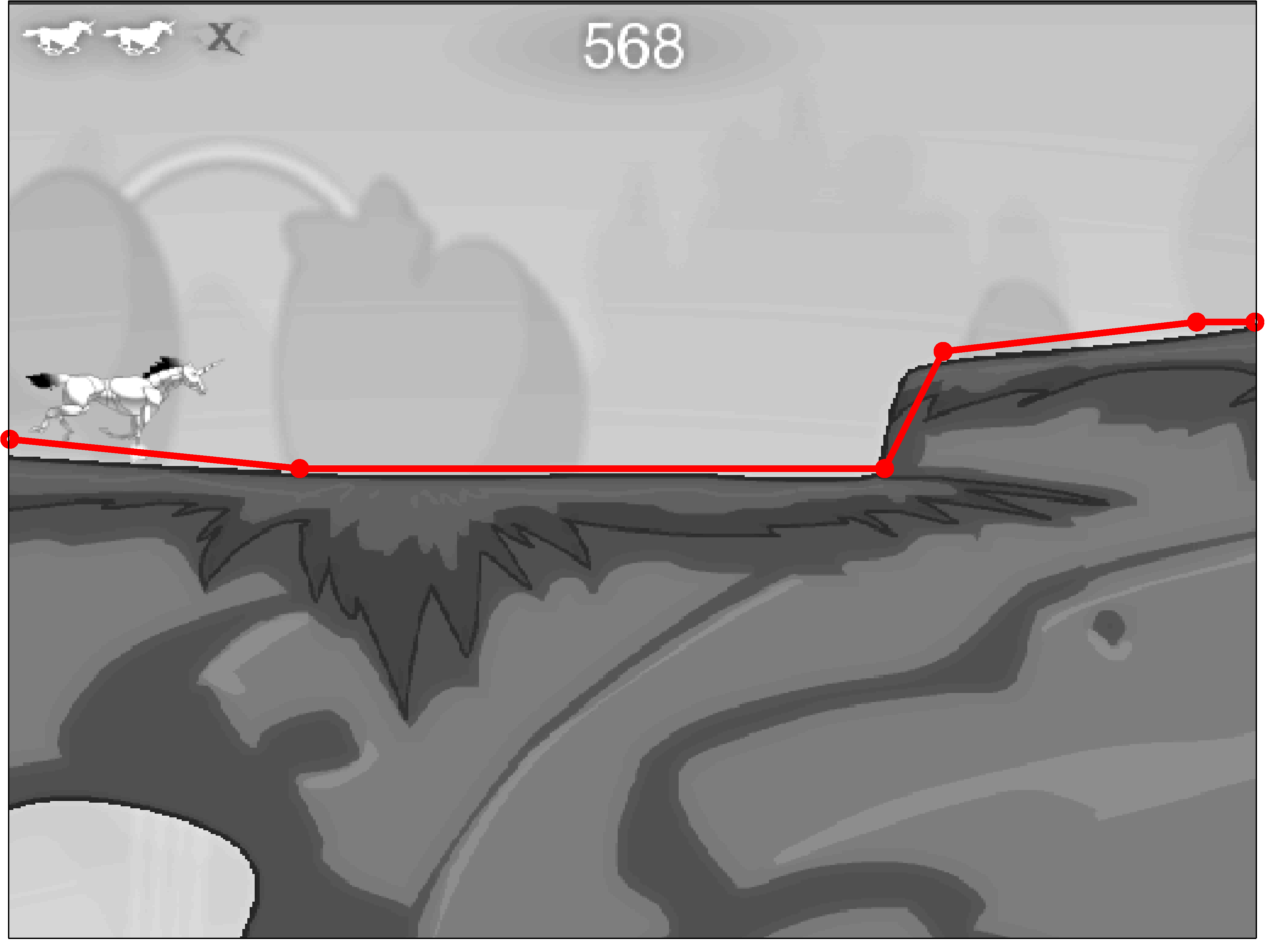}
\par\end{centering}
}
\par\end{centering}
\caption{{\footnotesize{}a) Screenshot from the Robot Unicorn Attack with dense
contour points (green dots) and a coarse approximation of the surface
after the application of Douglas-Peucker algorithm (red circles).
This coarse approximation is used to determine the top of the platform
(shown in b) as a red line).}}
\end{figure}

The only difference in visual processing required for Robot Unicorn
Attack is the addition of the curve approximation algorithm, since
the shape of platforms in this game is more complex than in Canabalt
(Figure \ref{fig:Curved-platform}). 
\begin{figure}
\noindent \begin{centering}
\subfloat{\includegraphics[scale=0.25]{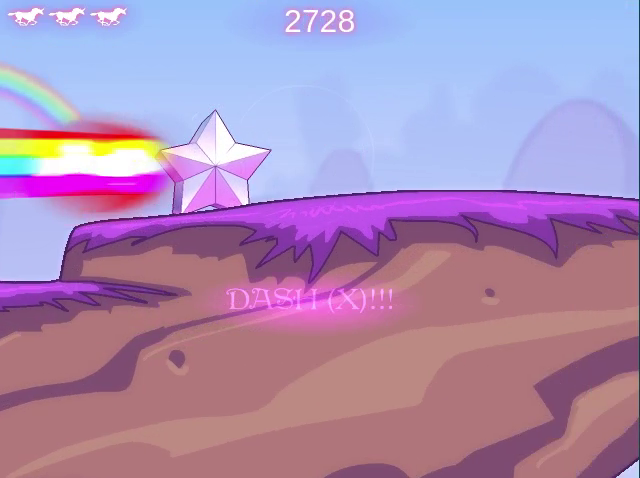}}\,\,\,\,\,\,\,\,\subfloat{\includegraphics[scale=0.25]{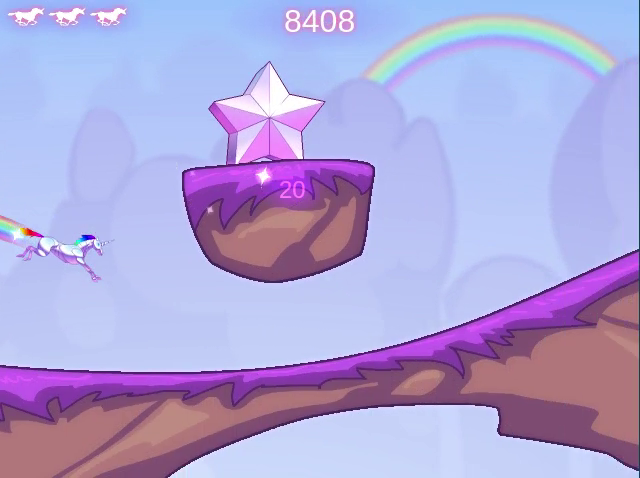}}
\par\end{centering}
\caption{\label{fig:curved_platforms_robot_unicorn-1}{\footnotesize{}Examples
of curved platforms in Robot Unicorn Attack}}
\end{figure}
A sketch of the curve tracing procedure in Cognitive Programs is outlined
in \citep{Tsotsos2014}, however, it was not possible to implement
it in real time within our system. In our implementation we first
threshold the image at $60\%$ of the intensity to eliminate the background,
resize the image to $64\times64$ pixels and compute connected components
as described in \citep{Wu2009}. For all blobs with area of $>20$
pixels we find contour points\footnote{Using the open source library OpenCVBlobsLib \url{http://opencvblobslib.github.io/opencvblobslib/}.}.
Next, the Douglas-Peucker algorithm \citep{DOUGLAS1973} is applied
to the contours to obtain a coarse polygonal approximation of the
platforms. This reduces the set of points by a factor of 15 (Figure
\ref{fig:unicorn_DP}). The top of the platform is found by tracing
a path from the leftmost to the rightmost point in the contour set
(Figure \ref{fig:unicorn_platform_top}). Finally, all coordinates
are rescaled to the original window size of $512\times512$ pixels.
For the classification step we follow the same steps as in Canabalt.
CNN parameters remained the same and the network was retrained on
the new set of patches representing objects in the Robot Unicorn Attack
(Figure \ref{fig:unicorn_objects}). The bandwidth for mean shift
is increased to $50$ to fit larger objects in the game.

We apply a similar technique to learn jump parameters for Robot Unicorn
Attack. Unlike Canabalt, it is rendered in a square window and since
the lookahead is much smaller, most of the jumps are made when the
next platform is not visible. To alleviate this, the game allows the
character to float in the air if the jump key is pressed continuously.
Our algorithm presses the jump button when the unicorn is close to
the edge of the platform and holds it until the next platform appears
in the field of view. Once it is visible again, it uses the pre-learned
jump parameters to estimate whether the platform can be reached if
the key is released immediately.

The strategy for playing Robot Unicorn Attack is different too since
final score depends both on the distance traveled and bonus points
objects awarded for picking objects (fairies). Simply staying alive
brings approximately 1000 points per 10 seconds of gameplay. Bonus
for destroying the stars starts at 100 and destroying every next star
increments this amount by 100 points. If the sequence is broken, the
bonus is reset to the initial 100 points. For fairies the initial
bonus (and increment) is 10. However, trying to collect all stars
and fairies is a risky strategy since often stars are placed on the
platform below or above the unicorn and are not easily accessible.
The easier and safer strategy is to collect most of the fairies, which
are usually placed along the safest trajectory between the two platforms.
When the star is on the same platform as unicorn, it is destroyed,
otherwise it is safer to ignore it. 

\section{Evaluation}

Playing video games has many subjective aspects, such as style, strategy
and entertainment value, which are hard to quantify, but are important
factors when judging one's performance. We also cannot evaluate STAR-RT
against other game AI algorithms. To the best of our knowledge, there
is no software that can play Canabalt or Robot Unicorn Attack, nor
can other existing game playing algorithms be easily adapted to this
task (mainly because current approaches require a simulator for training
and testing, which is not available for the games we use). Even though
the machine learning approaches such as DeepMind system \citep{Mnih2015}
may eventually learn to play these games, our goal for this paper
was to create a transparent model of human visual system and attention,
and to demonstrate the connections between its multiple components. 

Since each game session is randomly generated, it is not possible
to directly compare the performance of human players to our algorithm
with respect to jump precision and other elements of gameplay. Therefore,
we use the final score as our main evaluation metric and additionally
report the performance of the vision module. 

\subsubsection*{Canabalt}

Correct identification of obstacles is crucial for playing the game.
To test the vision module we took 5000 screenshots from several recorded
games and generated ground truth for locations of the objects in each
frame and true displacements between consecutive frames. The ground
truth marks the center of each object as its location. The runner
is correctly detected in $98\%$ of the frames (detection is successful
if it is within $\nicefrac{1}{4}$ of the runner's height from the
middle of the character). For crates and robot drills the detection
rate is also high at $97\%$, where detection is considered successful
if it is within the object contour. Most of the misdetections are
caused by sudden shakes of the game screen when a robotic drill falls
on the rooftop or a rocket passes in the background. 

An average game lasts 48 seconds (approximately 4000 frames). Most
of this time the player does not directly control the character, since
running on top of the platform and flying through the air after the
jump are done automatically by the game engine. As a result, only
$15-20\%$ of frames are critical for successfully playing the game.
In particular, the frames immediately before obstacles are needed
for speed estimation and pressing the key. In order to determine the
correctness of speed estimation, we gathered ground truth data on
5000 frames by overlaying pairs of consecutive frames, measuring the
exact displacement in pixels, and computing the speed of scrolling
from these measurements. Since displacements between the frames are
not computed if the platform spans the whole screen, in that case
speed remains fixed at the last estimated value. We take this into
account in evaluation. The measured speed for the 100 frames preceding
the decision point was within $\pm50$ pixels per second from the
ground truth, which is negligible, as it has no effect on jump precision. 

In order to evaluate the overall performance of STAR-RT we use the
recent statistics from the browser version of Canabalt available at
\url{kongregate.com}\footnote{We use only scores recorded from PC version of the game, since releases
on other platforms such as Mac OS and Android differ in input method
and have slightly different physics. }. The score in the game depends only on the distance traveled and
is measured in meters. The all-time highest score reported for the
game on \url{kongregate.com} is 30,760 m. The average score is reset
weekly, but is usually around 2500 m. It can be concluded from the
statistics that a score of at least 10,000 m is reasonable for an
expert player.

\begin{figure}
\noindent \begin{centering}
\includegraphics[scale=0.2]{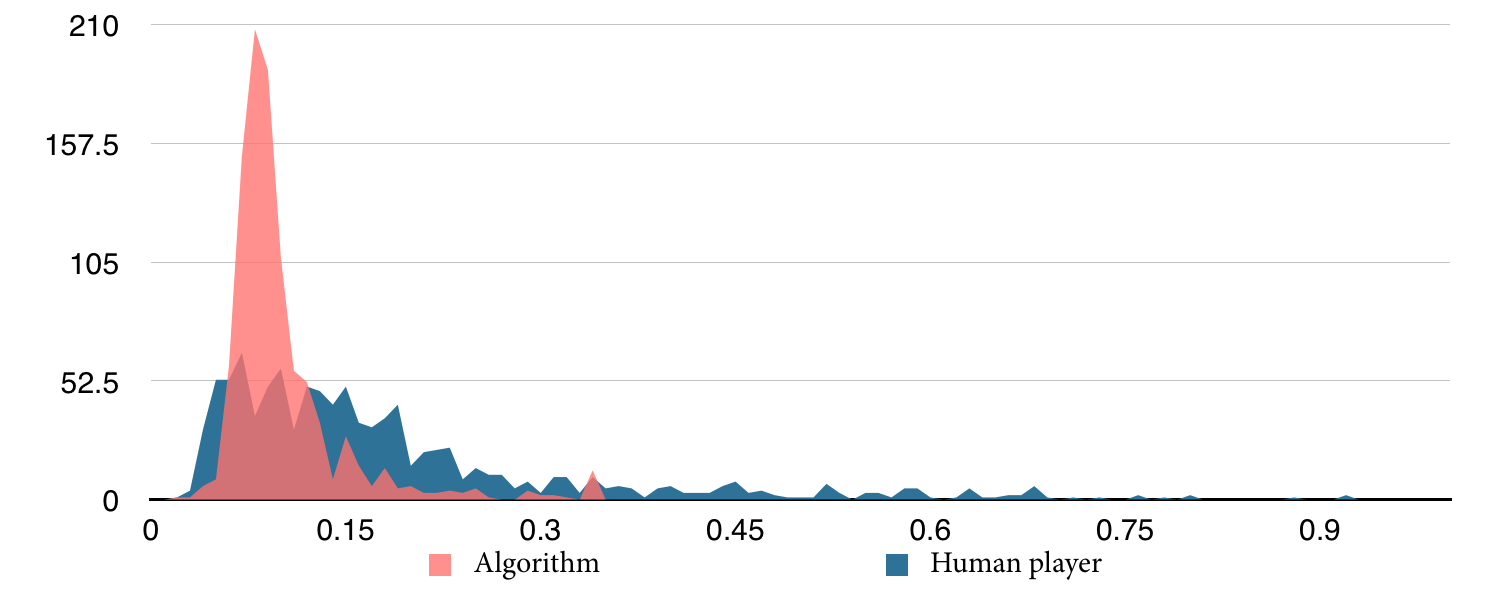}
\par\end{centering}
\caption{\label{fig:hist_button_press}{\footnotesize{}The histogram of key
press durations based on 100 games recorded with a human player and
STAR-RT playing.}}
\end{figure}

We collected the final scores of 1000 games played by STAR-RT after
the training was finished. The mean score was over 3000m and the top
score was 25,254 m, making it \#18 in the all-time best ranking posted
on \url{kongregate.com}. The most common reason for failing was hitting
a wall due to the bad jumping trajectory or timing. As mentioned before,
jumping is the essential skill in this game. The best strategy is
to land the character as close as possible to the edge of the platform
since it gives the player more time to make the next decision. However,
motor skills play a great role in executing this intention, especially
as the game reaches higher speeds where every millisecond counts.
This is demonstrated by the superior performance of STAR-RT, which
is tuned to jump as soon as possible and select the trajectory to
land close to the edge of the rooftop and has a better control over
the keyboard than a human player. Figure \ref{fig:hist_button_press}
shows the distributions of key press times for the human player (the
first author) and the algorithm: on a physical keyboard the average
press time is 180 ms, while the mean for the algorithm is 92 ms.

\subsubsection*{Robot Unicorn Attack}

Robot Unicorn Attack has more colorful graphics but visually it is
simpler than Canabalt: the camera movement is smooth, there are only
3 types of objects distinct from the background and fewer distractors.
The detection accuracy for the unicorn is at 95\% based on 5000 frames
collected from several recorded games (successful detections are within
$\nicefrac{1}{4}$ of unicorn's height from the center of the unicorn\textquoteright s
torso). The misdetections mostly happen when the unicorn is dashing
and temporarily disappears behind the explosion, however, it does
not affect the performance, since the controls are inactive at this
time. Speed estimation is easier in this game because many points
are generated for each platform. As a result, speed estimation is
within $\pm20$ pixels from the ground truth (generated by the same
procedure as for Canabalt). But in Robot Unicorn Attack platforms
are usually farther apart and the camera moves more than in Canabalt.
As a result, frequently no platforms are visible in the frame and
speed estimation is not possible. 

Although Robot Unicorn Attack was played over 32 million times on
the popular gaming site \url{adultswim.com} since its release in
2010, highest scores posted online are often fake. Due to the game's
popularity many ways of tampering with the game are published online.
Since many users submitted final scores of several billions of points,
the official scoreboard lost its meaning and was removed from the
site. However, based on the videos of expert players posted online
and our own experience we estimate that top players should achieve
around 100,000 points per run and close to 300,000 per game. Expert
players score approximately 80,000-100,000 points per game. 

The highest score achieved by STAR-RT is 26,591 points in a single
run and 72,697 points per game. We believe that the results can be
further improved by fully utilizing the control mechanisms in the
game, e.g., a double jump (i.e. jumping again in the air to correct
the trajectory).

By adding a second game of similar genre but visually very different
we were able to test the flexibility of Cognitive Programs. Few changes
were required in order to make the system play a new game. Namely,
the addition of visual processing routines to handle curved platforms,
retraining CNN for different types of objects, and changes in the
control function to add an extra key for dashing. Overall, the algorithm\textquoteright s
style of playing both games is markedly different from human players.
It makes more precise jumps than would be possible by using the physical
keyboard, has better reaction time and is able to perform consecutive
actions in quick succession.
\begin{table}
\noindent \begin{centering}
\begin{tabular}{c|c|c}
Image loading & 3.39 ms & 28\%\tabularnewline
\hline 
CPU time & 1.2 ms & 10\%\tabularnewline
\hline 
Total GPU time & 5.6 ms & 47\%\tabularnewline
\hline 
GPU overhead & 1.8 ms & 15\%\tabularnewline
\hline 
Total & 11.84 ms & \tabularnewline
\end{tabular}
\par\end{centering}
\caption{{\footnotesize{}\label{tab:Processing-times}Processing times per
frame}}
\end{table}

\subsubsection*{Parameters}

In STAR-RT 13 parameters tune various aspects of the system, which
include:
\begin{itemize}
\item 3 parameters for the foveation set to reflect the experimental setup
we used for the human players: fovea radius (set to 2 degrees of visual
angle), resolution of the monitor (92 ppi) and the distance from the
monitor (57 cm).
\item 7 parameters for the visual hierarchy and WM: number and length of
detected lines (10 and 100 px respectively), number of salient points
found with AIM (100), threshold for AIM saliency (> 50\%), number
and size of patches for classification (50 and 30 \texttimes{} 30
px), bandwidth of the mean-shift for finding object centroids (20
px).
\item 3 parameters for external functions: number of previous frames for
speed estimation (set to 15) and 2 parameters for jump parabola learned
from the data.
\end{itemize}
It should be noted that our system is robust to changes in most of
these parameters. The main effects from changing the values are observed
in the running time of the algorithm and affect accuracy less. For
instance, in order to reach the best results in Canabalt, several
thousands of jump trajectories were sampled from the past runs of
the algorithm. However, even a few hundred samples were sufficient
to play the game for some time. In order to set other parameters,
STAR-RT was tested with a range of values and the ones that ensured
the required time performance were selected. For instance, classification
is one of the more computationally expensive steps, but the game can
be played with the number of samples ranging from 20 to 100, optimal
being 50. 

\subsubsection*{Setup and performance}

The software for STAR-RT is implemented in C. The kernels for visual
processing on GPU are written in OpenCL 1.2. A visual debugger is
written in OpenGL 4.3 and GLSL 4.2. STAR-RT runs in a single thread.
In addition, MATLAB scripts were written for training convolutional
networks, gathering and analyzing various game-related data (``Game
Over'' screenshots, jump trajectories, image patches, etc.). All
experiments were conducted on a desktop running on Ubuntu 12.04.5
LTS with Intel Core i7-3820 @ 3.60GHz CPU, 16 GB of RAM, two AMD FirePro
W7000 (Pitcairn XT GL) video cards (one for visualization and one
for computation).

It takes $\sim12$ ms to process each frame, and the average frame
rate is approximately $84$ fps. The time for processing each frame
is not fixed and depends on the visual complexity of the scene, hence
frame rate fluctuates between 70 and 90 fps. Table \ref{tab:Processing-times}
shows the processing time for each major part of STAR-RT. Taking a
screenshot of the browser window takes $\nicefrac{1}{3}$ of the total
time. Other gameplay related operations such as decision making, finding
local maxima and mean-shift take negligible amount of time. The GPU
overhead is estimated to be $\sim15\%$ of the overall processing
time. 

\section{Conclusions}

The main contribution of this work is a positive test of Cognitive
Programs and STAR concepts in the domain of video games, opening the
road for more extensive implementations with the ultimate goal of
creating a general purpose visual system that can execute arbitrary
tasks formulated in natural language. STAR theory postulates that
this can be accomplished by decoupling the visual system from the
task control. To this end, the attention executive uses scripts, called
Cognitive Programs, to tune and guide the visual hierarchy represented
by the Selective Tuning model of visual attention. CPs share many
similarities with the earlier concept of Visual Routines, such as
decomposition of visual tasks into a set of context-independent atomic
operations and various attentional mechanisms. While past applications
of visual routines proved usefulness of this concept in areas from
robotics to computer vision and game AI, many of those works omitted
low-level vision and addressed other non-vision related issues such
as developing complex game playing strategies. By choosing games with
relatively simple game logic but visually complex appearance for testing
STAR-RT, we were able to focus on low- and intermediate-level vision,
gaze control and integrating visual data with symbolic reasoning.
In particular, for our proof-of-concept of STAR framework, we formulate
CPs for playing a game. These consists of multiple tasks, such as
finding a character, locating the platforms, etc, which in turn can
be split into more primitive subtasks, e.g. finding straight lines,
computing sparse optical flow and reading from and writing to memory.
Then we apply these CPs to another game with similar logic but very
different visual properties. Since both games (Canabalt and Robot
Unicorn Attack) are closed-source, they cannot be artificially slowed
down or dissected to find the internal parameters, which makes the
task more realistic.

Visual attention plays a major role in STAR framework. It includes
more than a dozen of different attentional mechanisms \citep{Tsotsos2011}
that have ample psychological and neurophysiological support. STAR-RT
successfully applied a subset of these, including foveation, gaze
control, spatial priming, bottom-up and top-down saliency and inhibition
of return. In fact, these mechanisms are essential for real-time performance
which puts an upper bound on a computation of $<20$ ms per frame. 

In order to satisfy the real-time constraints, the original concept
of visual hierarchy (as the most computationally expensive) had to
be modified, even with most of the processing moved on GPU. For instance,
the only existing implementation of ST takes 1 second to process a
single low-resolution image ($256\times256$ px). The simplified version
of the VH used in STAR-RT is much faster but cannot be efficient without
multiple attention mechanisms. For instance, in STAR-RT both foveation
and the bottom-up AIM saliency allow reducing the search space to
a few salient regions. Together, the saliency map and FSM provide
the bottom-up and top-down guidance respectively for the next fixations,
while inhibiting attended locations in the saliency map prevents the
system from cycling between the most salient locations. 

Even though the STAR framework is not yet completely realized, parts
of it, such as fixation control, visual hierarchy (represented by
ST) and visual attention executive, have been previously implemented
and tested in isolation with the stimuli typically used for psychophysical
experiments. We put together major elements of STAR, some in simplified
form, to perform a more realistic task of playing a computer game.
By that we demonstrated that the STAR framework is relevant for performing
this and similar tasks. This work also identified the issues that
must be solved by any real-time cognitively plausible implementation
and uncovered related theoretical and engineering challenges, which
will inform future efforts to extending STAR to a broader range of
tasks. Some of these issues are discussed below.

While the idea of modular vision routines is supported by neurophysiological
and physiological studies, the nature and number of the specific low-level
vision operators remain an open research question. The problem of
decomposing a more abstract task into subtasks and passing it to the
visual system for execution is still relatively unstudied. To date,
providing task guidance for visual processing has been limited to
relatively simple cases such as using a class template or label to
bias localization (i.e. tasks like \textquotedblleft find object of
class X in the image\textquotedblright ) \citep{zhang2016top,zhou2016learning}.
In our work we attempted to represent a more complex task of playing
the game as a series of generic vision processing steps (e.g. edge
detection/curve tracing, saliency, object recognition, etc.), which
could be applied to other visual tasks as well. For example, in \citep{Tsotsos1992}
the first step proposed for understanding an arbitrary image is two-dimensional
grouping of intensity-location values such as finding edges, regions
and flow vectors.

Finding the right data structures for task description and passing
messages within the framework, to the best of our knowledge, is an
open research problem and hard-coding the communication channels remains
the only viable solution for any relatively complex application. Learning
new representations, tuning them and combining them depending on the
task presents yet another set of unresolved issues. In particular,
past research on visual routines demonstrates that learning a sequence
of actions is a hard problem even in simple cases with only a few
parameters. This is one of the concerns for STAR, where every method
can be parametrized during runtime to adapt to changes in the environment
and task requirements. 

Since STAR was originally designed to work with static images, one
of the initial challenges was adapting the framework to dynamic stimuli.
In our implementation each frame is processed individually and then
candidates for objects and platform locations are recorded in the
vWM and tWM. External functions are called to compute displacements
between consecutive frames, match objects, and discard false positives.
While our solution was a compromise, it shows the functional importance
of working memory which is also prominent in the human visual system.
In the future, a more biologically plausible solution for moving stimuli
is needed.

Currently the STAR diagram shows only static connections between elements
and directions of the information flow, some of these processes should
run asynchronously to properly mimic human visual system. STAR-RT
runs on a single thread and avoids this problem by design, but any
future implementations would have to address the synchronization of
recurrent processes.

\begin{figure}[H]
\noindent \begin{centering}
\includegraphics[scale=0.3]{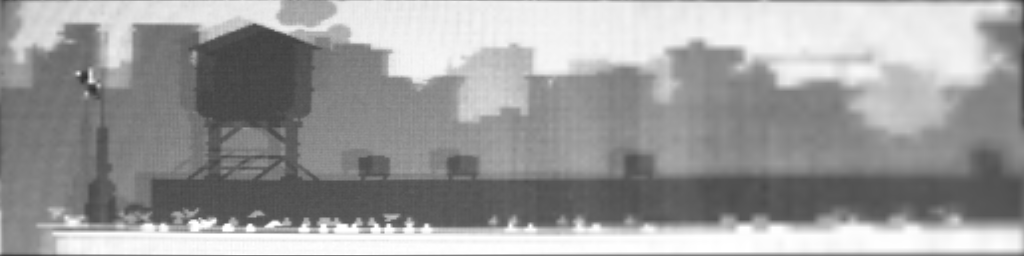}
\par\end{centering}
\caption{{\footnotesize{}\label{fig:camera_image}Image of the game screen
taken with a Point Grey Flea3 camera (foveation is added at post-processing).
Because of the lens distortion the rooftop surface is slightly bowed
inwards. Ghosting is also visible as a replica of the previous frame
superimposed on top of the current frame. In addition, a Moiré interference
pattern can be seen in the image.}}
\end{figure}

To date, many of the implementations of Visual Routines and our implementation
of Cognitive Programs focused primarily on the synthetic images. However,
for many visual tasks a physical video camera might be required, which
would introduce an additional challenge. For instance, consider the
task of playing a game such as Canabalt with a real camera feed used
as input instead of screenshots. Our preliminary experiments show
that camera images are much harder to work with because of the ghosting,
motion blur, lens distortions and interference patterns introduced
by the monitor. These artifacts can be clearly seen in the Figure
\ref{fig:camera_image}. In particular, significant motion blur is
present even when filming at the high frame rate of $60$ fps. 

In conclusion, STAR-RT implements all major components of STAR such
as fixation control, task working memory, visual working memory with
the blackboard, relevant elements of the visual attention executive
(excluding the parts that interact with the full visual hierarchy)
and a reasonable approximation of visual hierarchy. These components
are applied to playing two closed-source games in real time using
only visual information that would be available to a human player.
Although STAR-RT demonstrates some similarities to human player behavior
by design (e.g. fixation patterns), they may be hard to quantify. 

Since all of the components of the system are tightly interconnected,
it is not possible to evaluate precisely how each of them affects
the performance of STAR-RT as the removal of any part of the system
would lead to the system not being able to play the game. For instance,
if AIM saliency is removed, then many more samples would be passed
to the object classification stage, which in turn would increase the
processing time and add extra noise due to the classification errors. 

The possible applications for STAR-RT can be extended beyond what
was presented in this paper. For example, a system based on human
vision and provided with the same information as a human player, might
be useful to imitate human player performance. This topic has recently
been of interest to the game AI community. Such system would also
make more modern games available for experimentation, since most of
AI research has been done on emulators of outdated console games.
Finally, a framework for performing complex visual tasks has a direct
application in mobile robotics, especially for tasks that involve
active vision and interaction with the environment in real time.

\bigskip{}

\subsubsection*{Acknowledgment}

This research was supported by several sources, through grants to
the second author, for which all the authors are grateful: Air Force
Office of Scientific Research (FA9550-14-1-0393), Office of Naval
Research (N00178-15-P-4873), the Canada Research Chairs Program (950-219525),
and the Natural Sciences and Engineering Research Council of Canada
(RGPIN/4557-2011).

\bibliographystyle{elsarticle-harv}
\addcontentsline{toc}{section}{\refname}\bibliography{msc_bibliography}

\end{document}